\documentclass{egpubl}
\usepackage{pg2026s}

\WsConferencePaper

\usepackage[T1]{fontenc}
\usepackage{dfadobe}  

\usepackage{cite}  
\BibtexOrBiblatex
\electronicVersion
\PrintedOrElectronic
\ifpdf \usepackage[pdftex]{graphicx} \pdfcompresslevel=9
\else \usepackage[dvips]{graphicx} \fi

\usepackage{egweblnk} 
\usepackage{amssymb}
\usepackage{amsmath}
\usepackage{microtype}
\usepackage{mathtools}
\usepackage{listings}
\usepackage{xcolor}
\usepackage{algorithm}
\usepackage{algpseudocode}
\usepackage{subcaption}
\usepackage{booktabs}
\usepackage{multicol}
\usepackage{multirow}
\usepackage{marginnote}
\usepackage[normalem]{ulem}
\usepackage[nameinlink,capitalise]{cleveref}
\usepackage{caption}

\def\ie{\emph{i.e.}}
\def\eg{\emph{e.g.}}

\newcommand{\xmark}{$\times$}

\newcommand{\ourbench}{\textcolor{black}{our dataset}}
\newcommand{\Ourbench}{\textcolor{black}{Our dataset}}
\definecolor{highlight}{HTML}{000000}
\newcommand{\highlight}[1]{\textcolor{highlight}{#1}}
\title{TruncGradGS: Improved 3D Gaussian Splatting via\\ Truncated Gradient Updates}

\author[T. Morales, N.-Q. Le-Pham, R. Atkins \& B.-S. Hua]
{\parbox{\textwidth}{\centering
Théo Morales$^{1}$\orcid{0000-0002-2275-0895}
\quad
Nhat-Quynh Le-Pham$^{1}$\orcid{0000-0002-8668-9691}
\quad
Robin Atkins$^{2}$
\quad
Binh-Son Hua$^{1}$\orcid{0000-0002-5706-8634}
\\[2mm]
{\parbox{\textwidth}{\centering
$^1$Trinity College Dublin \quad $^2$Dolby Laboratories
}
}}
}

\begin{document}


\maketitle
\begin{abstract}
   
   3D Gaussian Splatting has become a de facto scene representation for novel
   view synthesis, yet robustly learning 3D Gaussian primitives from visual
   input remains challenging. Standard optimization relies on gradient-based
   updates, but a common issue is the gradient vanishing phenomenon: a pixel far
   from a Gaussian primitive often has diminishing gradient magnitudes to
   influence primitive attributes, resulting in suboptimal scene reconstruction.
    In this paper, we propose a method to address gradient vanishing with a
    piecewise truncated gradient formulation that improves the optimization
    stability and robustness to initializations. We show that our method
    consistently improves 3D Gaussian Splatting with random and COLMAP
    initializations while being generalizable across static and dynamic Gaussian
    Splatting. As a by-product, we also examine the limitations of current
    benchmarks for dynamic scenes, and introduce a novel dataset for benchmarking
    dynamic Gaussian Splatting using synthetic 3D scenes. We demonstrate the effectiveness of our method in both static and dynamic settings for the public benchmarks and our proposed dataset.

\ccsdesc[300]{Computing methodologies~Rasterization}
\ccsdesc[500]{Computing methodologies~Reconstruction}
\printccsdesc   

\end{abstract}  

\begin{figure}[tb]
    \centering
    \includegraphics[width=\linewidth]{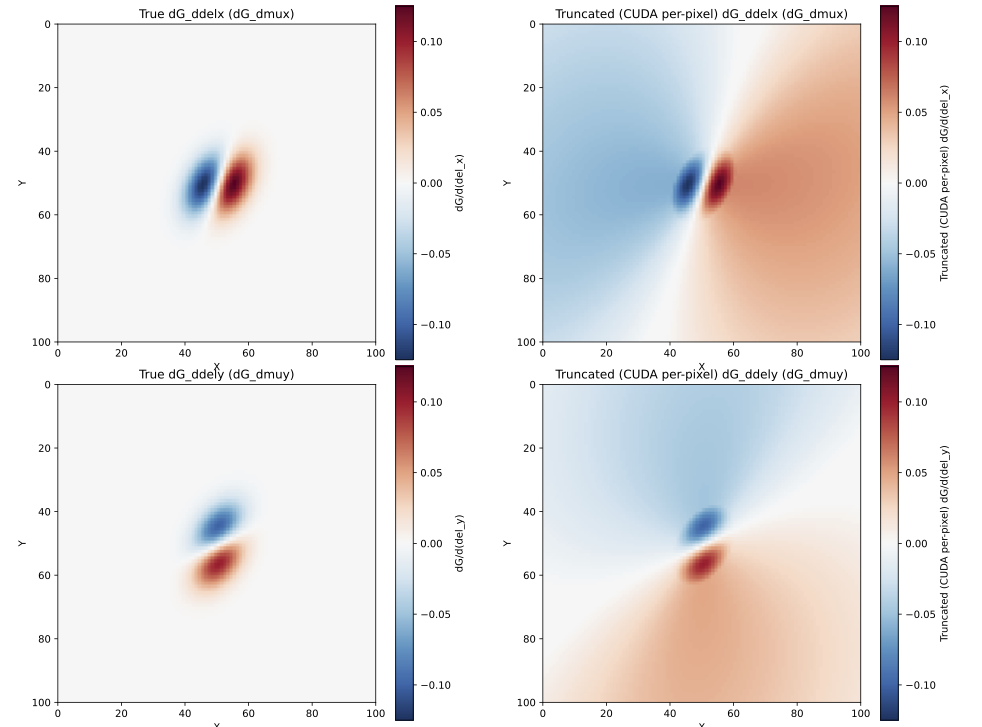}
    \vspace{1mm}
    \caption{Visualization of our piecewise truncated gradient field. The
    example assumes a 2D Gaussian primitive that splats at the center of an
    image. The left column shows the derivative of the Gaussian distribution
    w.r.t. its mean on the $x$ and $y$ axis. It can be seen that the
    gradient is near zero for any pixel outside the local support. In the right
    column, the piecewise gradient fields for the same Gaussian
    splat show a wider support with stronger gradient signals
    in far-away regions. Our piecewise gradient field is composed of the true
    Gaussian derivative inside the isocontour of the Gaussian splat, and of a
    linear surrogate outside. We define the isocontour by a density threshold and ensure gradient continuity at the contour boundary.
    } \label{fig:truncated-grad}
\end{figure}

\section{Introduction}
3D Gaussian Splatting has emerged as a de facto scene representation in visual computing, supporting diverse applications in novel view synthesis, scene reconstruction, and content creation. 
Its central idea is to model the radiance field of a scene using a collection of
ellipsoids formulated as 3D Gaussian primitives. Each 3D Gaussian is
parametrized by a mean, covariance matrix, and appearance-related attributes
such as opacity and spherical harmonics coefficients. These primitives are
typically learned through optimization. However, robustly learning such Gaussian
primitives remains challenging, particularly in the setting of random initialization
or dynamic scenes.

When fitting a scene, the set of 3D Gaussian primitives is learned from image observations.
The optimization is uses a differentiable rendering pipeline, where the Gaussian
primitives are rendered via camera projection and primitive rasterization. This
pipeline is fully differentiable, enabling the optimization of the Gaussian
attributes via stochastic gradient descent. To parallelize rendering, the image
plane is tiled into pixel patches (e.g., $16 \times 16$), allowing the dispatch
of a thread block for handling the rasterization using one thread per pixel. To
efficiently process large primitive sets, optimizations are performed locally:
only  Gaussians that splat onto a tile are considered for receiving gradients
for attribute updates from that tile, effectively reducing the optimization
workload per tile. 

The original tiling scheme, while enabling resource-efficient training and
inference, can reduce learning stability and effectiveness. As a Gaussian
primitive only contributes within tile locality, it receives no gradient signals
from pixels outside its support, and therefore cannot be used to reconstruct
those pixels. 
As a result, reconstruction quality might degrade when a Gaussian primitive fail
to move toward distant yet relevant pixel regions. To mitigate this issue, one
can use adaptive density control to improve scene coverage, which clones and
splits Gaussian primitives with perturbed positions so that new Gaussians can
progressively explore the scene further from their initializations.
%
One can also expand the radius support for a Gaussian primitive as a simple way of
collecting gradients from pixels further away. Nevertheless, the Gaussian
derivative remains near zero at the tails of the Gaussian distributions, preventing
far-away pixels from providing meaningful updates~\cite{zhang_papr_2023}.
While such limitations were addressed in prior work~\cite{zhang_papr_2023,
NEURIPS2024_93be245f}, there remains a need for a general approach that applies
across scenarios such as  random/COLMAP initializations and static/dynamic scene settings. 

To overcome the limitations of 3D Gaussian rasterization, we propose to modify the Gaussian
derivatives to strengthen the primitive/tile mutual coverage while keeping the
effectiveness of the current tiling scheme. Particularly, we derive a piecewise
truncated variant of the gradients by using a linear function to expand the
tails of the Gaussian derivatives, and at the same time, maintaining the
derivative continuity (see \cref{fig:truncated-grad}).
We show that such truncated gradients can be integrated into the learning process and improving the scene quality across initializations. 
More importantly, we show that truncated gradients work well for both vanilla Gaussian Splatting and dynamic Gaussian Splatting. 
We provide extensive experiments on static and dynamic scene datasets. 
Additionally, while recent works in dynamic Gaussian Splatting show impressive results on the
available benchmarks, their performance on more challenging dynamic scenes
remains untested. To address this gap, we propose a novel dataset, introducing
six realistic synthetic scenes rendered into multi-view videos, showcasing complex
dynamics in motion and appearance. 
Our benchmark of the state-of-the-art methods in dynamic Gaussian Splatting shows that our method  outperforms baselines on these challenging
motion, atmospherics, and geometry examples.  We envision that such additional data could motivate future research on dynamic Gaussian Splatting with potentially significant improvement. 
We will make our code and data available upon publication.

In summary, our contributions are as follows:
\begin{itemize}
    \item We provide a formal analysis of the partial derivative
    of 2D Gaussians that shine light on the vanishing gradient problem. While this is sometimes
    discussed in the literature, this work pin-points the root of the problem on
    the theory and implementation of 3DGS.
    \item We present a piecewise truncated gradient field to address the
    vanishing gradient problem of 3DGS. Our modification effectively widens the
    support of 2D Gaussian splats. We apply this modified gradient during
    training by expanding the radius of the splats to extend their tile
    coverage.
    \item We demonstrate the effectiveness of truncated gradients on both static
    and dynamic settings. In the static setting, we show consistent improvement
    for random and COLMAP initializations. For the dynamic setting, we show
    consistent improvement on SOTA methods.
    \item We introduce a novel dataset of six synthetic scenes for multi-view 4D
    reconstruction. This presents a new benchmark for raising the quality standard for dynamic 3D
    Gaussian Splatting methods.
\end{itemize}

\section{Related Works}

\paragraph*{Robust optimization.}
The original 3D Gaussian Splatting (3DGS)~\cite{3dgs} heavily relies on Structure-from-Motion (SfM) point clouds for initialization and heuristic adaptive density control (ADC) for densification/pruning. This often leads to sensitivity to poor or sparse initialization, vanishing gradients in under-constrained regions, floaters, and inefficient Gaussian proliferation, especially in sparse-view or long-range fitting scenarios.

Several recent works target the initialization bottleneck. RAIN-GS~\cite{jung2024raings} relaxes the dependency on accurate SfM points through sparse-large-variance (SLV) initialization, progressive low-pass filtering, and adaptive bound-expanding splits, enabling robust training even from random point clouds. Librated-GS~\cite{pan2025liberated} further eliminates SfM reliance via monocular depth alignment, progressive segmented initialization, and importance-aware resampling. Other approaches explore stochastic/MCMC-based optimization that reframes densification as probabilistic sampling for improved robustness~\cite{kheradmand2024mcmc}. Additional SfM-free strategies include hierarchical training with video frame interpolation~\cite{ji2025sfgs}, dense feature matching to improve point cloud initialization in sparse-view scenarios~\cite{seibt2024dense}, and various depth-prior initialization pipelines~\cite{desiatov2026role}.

For densification and pruning, methods have moved beyond naive gradient-based cloning/splitting. Taming 3DGS~\cite{mallick2024taming} introduces guided, purely constructive densification with importance scoring and strict budgeting for controlled Gaussian counts. ConeGS~\cite{baranowski_conegs_2025} uses error-guided densification along pixel viewing cones with a geometric proxy, achieving superior quality with fewer primitives. Further advances include Pixel-GS for pixel-aware gradient control \cite{pixelgs}, AD-GS with alternating high/low densification phases for sparse inputs~\cite{patle2025adgs}, complexity-density consistency strategies~\cite{dong2025reframing}, perceptual and generative densification~\cite{nam2025generative}, ESA-GS with elongation splitting and assimilation~\cite{chen2025esa}, reconstruction-aware adaptive pruning schedulers~\cite{wang2026prune}, learning-to-prune frameworks with Gumbel-Sigmoid masks (LP-3DGS)~\cite{zhang2024lp3dgs}, and significance-aware or spatio-spectral pruning~\cite{luo2026efficient}. These works are particularly relevant for random or sparse initialization settings as they reduce heuristic dependency, mitigate vanishing gradient issues, and improve stability in challenging geometries.

Recent works have also explored feed-forward or optimization-free Gaussian reconstruction pipelines that directly predict Gaussian primitives from sparse observations using neural networks~\cite{li2024lgm, szymanowicz2024flash3d, hong2025pf3plat}. Unlike optimization-based 3DGS methods that iteratively refine Gaussian attributes through differentiable rendering, these approaches amortize reconstruction into a learned inference process for significantly faster scene generation. While feed-forward approaches improve scalability and inference speed, they typically require large-scale training data and may generalize less robustly to out-of-distribution scenes or highly sparse viewpoints. In contrast, optimization-based methods remain more flexible and adaptable across diverse capture conditions, motivating continued research on improving optimization robustness and stability.

Compared to existing methods, our work takes a different perspective by directly addressing the locality and vanishing-gradient limitations of Gaussian optimization. Whereas prior approaches primarily focus on improving initialization, densification, pruning, or deformation modeling, we instead modify the gradient field itself to enlarge the effective optimization region of Gaussian primitives during training. The closest work to ours is PAPR~\cite{zhang_papr_2023}, which also addresses vanishing gradients in point-based rendering. However, PAPR focuses on point radiance fields and stochastic point propagation, whereas our method derives a piecewise truncated gradient formulation specifically designed for Gaussian splatting and tiled rasterization. Our approach is lightweight, easy to integrate into existing static and dynamic 3DGS frameworks, and complementary to existing initialization and densification strategies.

\paragraph*{Dynamic 3DGS.}

Recent works in dynamic Gaussian splatting model temporal dynamics either through spatio-temporal representations~\cite{yan_4dgs_2024, Wu_2024_CVPR, TiNeuVox, chen_dash_2025, xu_grid4d_2024, yang2024real} or heterogeneous deformation models~\cite{lee_fully_2024, splinegs, wu_localdygs_2025, katsumata2024compact, yeom2026trigs}. Despite the increasing complexity of these approaches, simpler linear velocity-based models remain among the most robust~\cite{4dgs, duan:2024:4drotorgs, disentangled_4d, wang_freetimegs_2025, dai20254dgv}.
Neural methods typically learn deformation fields using MLPs or 4D encodings~\cite{10657752, 4dgs, Cao2023HexPlaneAF, 10.5555/3737916.3741850, chen_dash_2025, oh2025hybrid}, while explicit approaches define motion analytically through polynomials, Taylor series, or B-splines~\cite{lee2024ex4dgs, hu2025learnable, splinegs}. Many works also adopt hybrid static-dynamic partitioning to reduce parameter count over long sequences~\cite{lee2024ex4dgs, oh2025hybrid, chen_dash_2025}.
One of the latest methods, 4D-Scaffold~\cite{cho2024scaffold}, proposes a memory-efficient 4D anchor-based framework with dynamic-aware anchor growing. It extends 3D scaffolding to the 4D domain using sparse grid-aligned anchors and compressed features, achieving high visual quality and fast rendering while greatly reducing storage costs for long sequences.

Despite of such advances, existing methods often struggle with large temporal divergence across frames. While adjacent frames show small motion, distant frames can differ drastically, leading to accumulated errors, mode collapse, and significant optimization difficulties. In particular, the vanishing gradient problem inherent in standard 3DGS rasterization becomes a critical bottleneck in dynamic settings, as rapidly moving or distant primitives receive negligible gradient signals.
To address this fundamental limitation, we apply our piecewise truncated gradient formulation (which replaces the near-zero Gaussian derivative tails with a linear surrogate) to dynamic Gaussian splatting. This modification widens the effective support of each Gaussian primitive, enabling more effective gradient flow even for large inter-frame displacements, while remaining fully compatible with existing dynamic frameworks.

The performance evaluation of dynamic 3DGS methods has been extensively studied using several public benchmarks. However, existing datasets often fall short in aspects critical for industrial applications. Many are limited to short sequences ~\cite{mildenhall2020nerf,technicolor,pidg}, lack sufficient photorealism ~\cite{mildenhall2020nerf, pidg}, or contain insufficiently challenging motion on which state-of-the-art methods have largely plateaued ~\cite{li2022neural, enerf, technicolor}.
\begin{figure*}[bt]
    \centering
    \includegraphics[width=\textwidth]{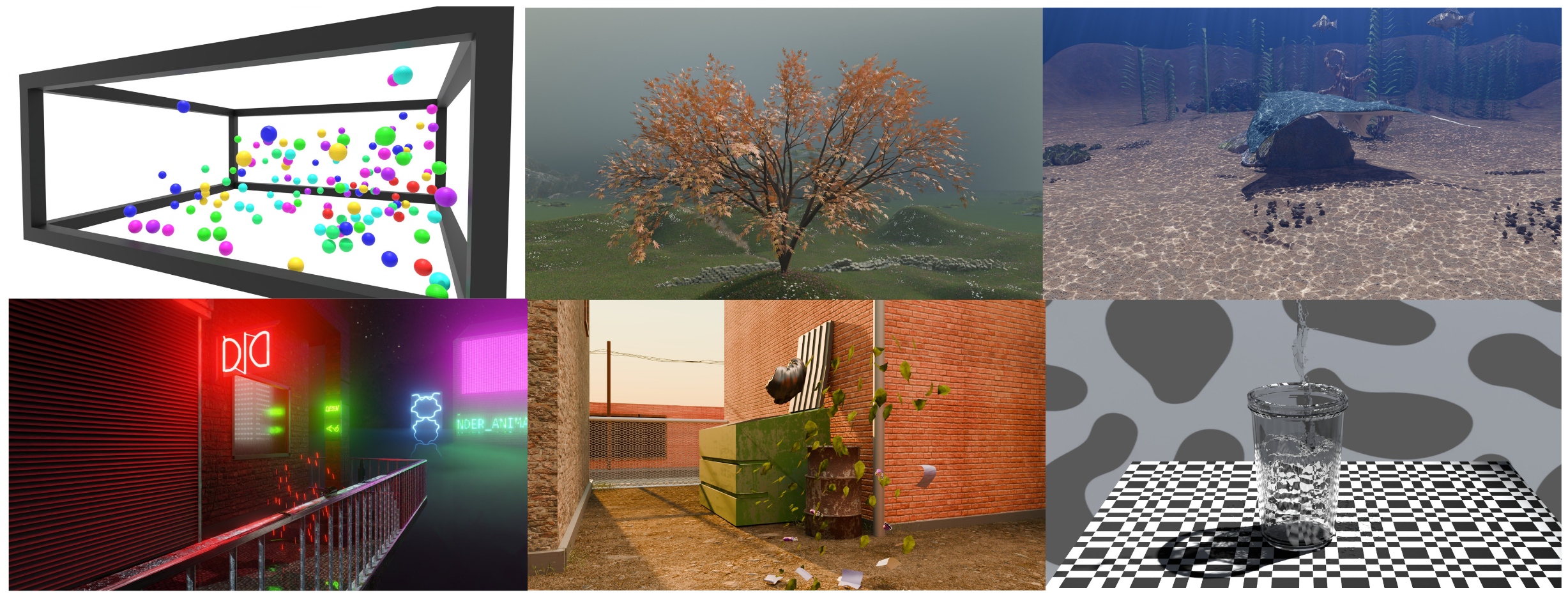}
    \caption{Still frames from our novel benchmark. We introduce six synthetic
    scenes, each as an animation of 300 frames (10s), ranging from $25$ to $50$ cameras
    arranged from $180$ to $360$ deg arrays. All scenes were either hand animated or
    entirely simulated in Blender. Our novel benchmark showcases complex geometry,
    fluid simulation, atmospherics, particles and caustics, significantly raising
    the bar for dynamic scene reconstruction methods, with 1 to 4 test view points per
    scene.}
    \label{fig:our_dataset_vis}
\end{figure*}
Our benchmark addresses these limitations by providing long-duration sequences with highly complex non-rigid motion, photorealistic quality, and diverse real-world scenes, making it more suitable for evaluating industrial-grade dynamic reconstruction methods.
We summarize the comparison in Table~\ref{tab:dynamic_benchmarks}.

\begin{table*}[bt]
    \centering
    \caption{\textbf{Comparison of public dynamic scene reconstruction benchmarks.} ``Complex motion'' refers to challenging non-rigid deformations; ``Long duration'' indicates sequences typically exceeding several hundred frames (suitable for evaluating long-term consistency); ``Realistic'' denotes photorealistic real-world captures versus synthetic data; ``Scene variety'' covers diversity in environments and objects; ``Multi-camera'' indicates synchronized multi-camera setups.}
    \label{tab:dynamic_benchmarks}
    {
    \begin{tabular}{l|ccccc}
    \toprule
    Benchmark & Complex motion & Long duration & Realistic & Scene variety & Multi-camera \\
    \midrule
        N3DV~\cite{li2022neural} & \xmark & \checkmark & \checkmark & \xmark & \checkmark \\
        Technicolor~\cite{technicolor} & \xmark & \xmark & \checkmark & \checkmark & \checkmark \\
        NeRF synthetic~\cite{mildenhall2020nerf} & \xmark & \xmark & \xmark & \checkmark & \xmark \\
        ENeRF-Outdoor~\cite{enerf} & \xmark & \checkmark & \checkmark & \checkmark & \checkmark \\
        HyperNeRF~\cite{park2021hypernerf} & \xmark & \xmark & \checkmark & \xmark & \xmark \\
        PIDG~\cite{pidg} & \checkmark & \xmark & \xmark & \checkmark & \xmark \\
        \textbf{Ours} & \checkmark & \checkmark & \checkmark & \checkmark & \checkmark \\
    \bottomrule
    \end{tabular}
    }
\end{table*}
\section{Methodology}
We begin with a brief introduction to the 3D Gaussian Splatting framework and an analysis of the vanishing gradient problem. This is followed by the introduction to the piecewise  truncated gradient, our main contribution.
\subsection{Background}

The standard setup of 3D Gaussian Splatting is to parametrize a radiance field
by a set of 3D Gaussian primitives with color and transparency attributes. The
radiance field is optimized to match a set of observed images. 
Each primitive is an ellipsoid represented by a Gaussian distribution
$G(x)=e^{-\frac{1}{2}(x-\mu)^T\Sigma^{-1}(x-\mu)}$ with mean $\mu \in
\mathbb{R}^3$ and covariance $\Sigma \in \mathbb{R}^{3\times3}$.
To model the appearance, these have an opacity parameter $\sigma \in \mathbb{R}$
and a view-dependent color $\mathbf{c} \in\mathbb{R}^3$, parameterized by
spherical harmonics. 
These parameters are optimized by minimizing the error
between the ground truth and the rendered image obtained from the Gaussian
representation.
The radiance field is rendered via rasterization, where
each pixel value $\mathbf{c}$ is computed by alpha blending the sorted set of
2D Gaussian splats $\mathcal{N}$ along a ray:
\begin{equation} \label{eq:alphablending}
    \mathbf{c} = \sum_{i\in\mathcal{N}} \mathbf{c}_i \sigma_i \alpha_i \prod_{j=1}^{i-1} (1-\sigma_i \alpha_j)
\end{equation}
where $\mathbf{c}_i$ and $\sigma_i$ the color and opacity of Gaussian $i$, respectively. 
$\alpha_j$ is the density value of the projected distribution from the corresponding 3D Gaussian density, obtained via splatting~\cite{ewa_splatting}: $\alpha_j =
e^{-\frac{1}{2}(x-\mu^\text{2D})^\top{\Sigma^\prime}^{-1}(x-\mu^\text{2D})}$ where
$x=[u~v]^\top$ is the 2D ray coordinates. The projected mean $\mu_{2D}$ is
obtained by projecting the 3D mean $\mu$ to the image plane with the camera
projection matrix. The projected covariance $\Sigma^\prime$ is defined by:
\begin{equation} \label{eq:splat_cov}
    \Sigma^\prime = JW\Sigma W^TJ^T
\end{equation}
where $J$ is the Jacobian of the affine approximation of the projective
transformation, and $W$ is the viewing transformation. The 2D covariance matrix
can be obtained by ignoring the last row and column of $\Sigma^\prime$, as
proven in~\cite{ewa_splatting}. 

To learn the scene parameters, the optimization algorithm starts with a sparse
initialization of the scene (either random or from Structure-from-Motion) and
then minimizes the reconstruction loss via Stochastic Gradient Descent (SGD).
The set of Gaussians is periodically densified and pruned by an adaptive density
control scheme, the details of which are explained in the original work~\cite{3dgs}.

In order to maintain a reasonable training time for a scene with dozens or
hundreds of images, the rasterizer makes two important decisions. Firstly, the
image is tiled in $16\times 16$ pixels to parallelize rasterization efficiently.
For each 2D splat, the largest of the two axes of its ellipse is used to compute
a bounding box which allows to allocate the primitive to all the image tiles it
intersects with. This effectively reduces the number of Gaussians to rasterize in each
CUDA kernel, at the cost of a limited support for the 2D splat, and hence less pixel error
contributions in the backward pass.
Secondly, for each pixel during rasterization, every Gaussian splat $i$ assigned
to this tile contributes some color by the scalar value $\sigma_i\alpha_i$, and
any Gaussian whose contribution scalar falls below a threshold is discarded. As
a result, the discarded Gaussians does not receive gradient updates for that
pixel.
We argue that those two key design choices are hindering the ability of 3D Gaussians
to explore the scene, which may result in a poor model for a poor scene initialization.
This is especially the case in dynamic methods (\eg,  4DGS~\cite{4dgs}), where distant
frames require different initializations and are challenging to model with the same set
of Gaussians. While relaxing these key constraints should, in theory, alleviate
these limitations, we show in the next section that the vanishing gradient problem
of the Gaussian distribution is the real optimization bottleneck.

\subsection{Gradient analysis}
During optimization, the Gaussian parameters are updated in a gradient backpropagation pass.
When the Gaussian splat obtained via perspective projection overlaps the pixels that it reconstructs,
all appearance attributes receive gradient updates. In the other case where the Gaussian does not splat
onto the pixels of interest, its appearance attributes cannot receive meaningful updates.
Therefore the spatial gradient, \ie $\frac{\partial L}{\partial\mu}$,
is critical to the success of the optimization, as the Gaussian primitives move and extend their coverage to match observed
images.
For this reason, we focus our analysis on the derivative of the optimization objective with
respect to the 2D mean in both $x$ and $y$ dimension.

When considering one of the 2D variables, e.g. $x$, the partial spatial derivative is
defined as
\begin{equation}\label{eq:dldmu}
    \frac{\partial L}{\partial \mu^{2D}_x} = \frac{\partial L}{\partial G^{2D}} \frac{\partial G^{2D}}{\partial \Delta_x} \frac{\partial \Delta_x}{\partial \mu^{2D}_x} 
\end{equation}
where $\Delta = (\mathbf{\mu}^{2D} - \mathbf{p})$ is the distance-space
parametrization of a pixel $\mathbf{p}$ for which we want to compute the 2D Gaussian density
\begin{equation}
    G^{2D}(\Delta) = 
        \exp{\{ -\frac{1}{2}\Delta^T{\Sigma^{\prime}}^{-1}\Delta \}},
\end{equation}
using the projected covariance $\Sigma^{\prime}$ of \cref{eq:splat_cov}.
The gradient of the loss with respect to the 2D Gaussian density is then defined as
\begin{equation}
    \frac{\partial L}{\partial G^{2D}} = \sigma \frac{\partial L}{\partial \alpha_i},
\end{equation}
which contains no interaction term with the Gaussian function.
However, the gradient of the Gaussian density with respect to the mean delta in $x$
and in $y$ are defined as
\begin{align}
    \frac{\partial}{\partial \Delta_x}G^{2D}(\Delta) &= -G(\Delta) \Delta_x a - G_i(\Delta) \Delta_y c,\label{eq:dgddelx} \\
    \frac{\partial}{\partial \Delta_y}G^{2D}(\Delta) &= -G(\Delta) \Delta_y b - G_i(\Delta) \Delta_x c, \label{eq:dgddely}
\end{align}
and clearly contain interaction terms with the Gaussian function $G^{2D}$. 
The coefficients $a, b, c$ are from the un-normalized general Gaussian 2D
function (conic form) used in the original formulation, expressed as
\begin{equation}\label{eq:conic}
    f(x,y) = \exp(-\frac{1}{2}[a(\mu_x-x)^2 + c(\mu_y-y)^2]-b(\mu_x-x)(\mu_y-y)).
\end{equation}
Since the Gaussian function has local support with an exponential falloff outside the mean, any pixel
outside the isocontour of the 2D splat receives near zero gradients due to the Gaussian
function zeroing-out equations \cref{eq:dgddelx} and \cref{eq:dgddely}, which in turns
nullifies the chain rule for \cref{eq:dldmu}.
We argue that the vanishing gradient of the Gaussian distribution is a strong
limitation for learning a scene configuration that diverges significantly from
the initializtion, as highlighted in previous research~\cite{zhang_papr_2023}. 
Since this naturally happens in dynamic scenes, we highlight the importance of
addressing this limitation in the 3DGS rasterizer.


\subsection{Piecewise truncated gradient}
\begin{figure}
    \centering
    \includegraphics[width=\columnwidth]{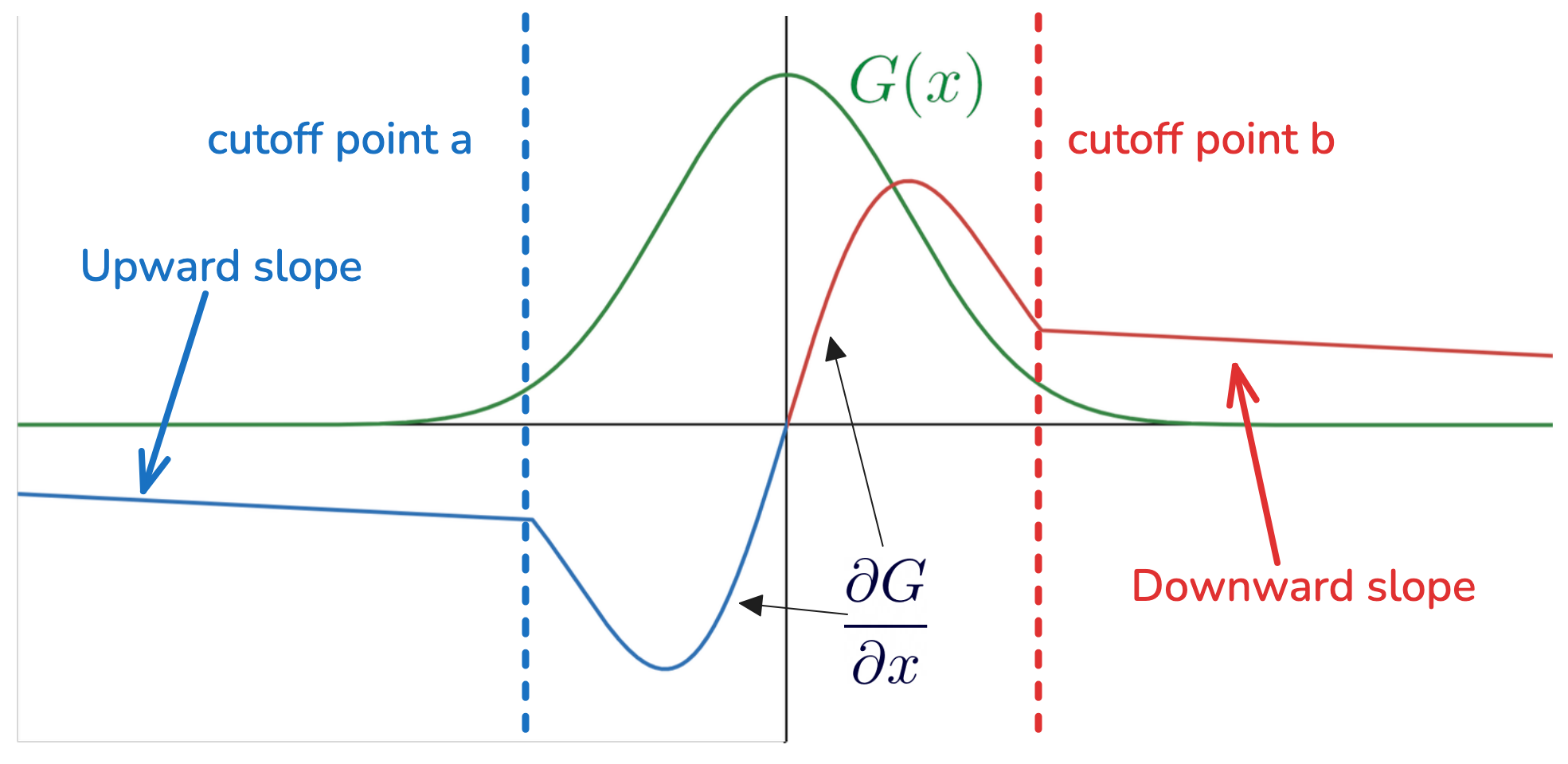}
    \caption{1D illustration of our piecewise gradient. Beyond the cut-off
    points, the partial derivative becomes a linear surrogate which converges to
    the true gradient. Between the cut-off points, we maintain the true partial
    derivative $\frac{\partial G}{\partial x}$.}
    \label{fig:piecewise_grad_1D}
\end{figure}
To remedy the vanishing gradient, we propose to truncate \cref{eq:dgddelx} and
\cref{eq:dgddely} such that the gradient changes to a linear surrogate below a
density threshold, as illustrated in \cref{fig:truncated-grad} and
\cref{fig:piecewise_grad_1D}, and is defined as the following piecewise gradient
field:

\begin{equation}\label{eq:dgdmu_truncated}
     \widetilde{\nabla G} = \begin{cases} 
         \frac{\partial G}{\partial \Delta_x} & \text{if } -2\log G < -2\log \tau \\ 
         \min(m\cdot x+b, \frac{\partial G}{\partial \Delta_x}) & \text{if } \frac{\partial }{\partial \Delta_x}G(b) < 0 \\
         \max(-m\cdot x+b, \frac{\partial G}{\partial \Delta_x}) & \text{if } \frac{\partial }{\partial \Delta_x}G(b) > 0
        \end{cases}
\end{equation}
with the Gaussian density threshold $\tau$ and the squared Mahalanobis distance
expressed in terms of the Gaussian function $G$:
\begin{align}
    d(p)^2 &= (\mu-p)^T \Sigma^{-1} (\mu-p) \label{eq:mahalanobis} \\
    &= -2 \log G \notag
\end{align}
The truncated gradient equals the true partial derivative 
when the squared Mahalanobis distance is less than a density threshold $\tau$, and follows a line equation otherwise. This line is defined by a slope $m$ set as a model parameter and a bias which we derive as explained below. \highlight{The slope $m$ controls the linear scaling of the gradient magnitude, effectively interpolating between the minimum-magnitude gradient and the gradient at the boundary point. The density threshold $\tau$ directly determines the boundary point and defines the isocontour of the 2D splat. We set $\tau$ to the alpha-blending discard threshold of the rasterizer.}

\paragraph*{Gradient continuity.}
Since our truncated gradient is piecewise linear and exponential, the only source of
discontinuity would be at the boundary between the two pieces. To ensure
continuity at the boundary, we match the linear surrogate $y=mx+b$ to the true
gradient at the boundary point $x_b$ -- the cut-off gradient :
\begin{equation}
   \frac{\partial }{\partial \Delta_x}G^{2D}(x_b) = mx_b+b,
\end{equation}
which yields 
\begin{equation}
    b = \frac{\partial}{\partial \Delta_x}G^{2D}(x_b) - mx_b.
\end{equation}
\highlight{The boundary point $x_b$ is defined via the Gaussian isocontour
$G^{2D}(x_b)=\tau$, or equivalently by substituting \cref{eq:mahalanobis}:
\begin{equation}
    G^{2D}(x_b)=\tau \equiv d(p)^2 = -2 \log \tau.
\end{equation}
To derive $x_b$, we parametrize it as the hit of ray with the Gaussian isocontour.
We define the ray as origin the query pixel
$p$ and unit direction $\frac{\mu - p}{||\mu-p||_2}$, such that 
\begin{equation} \label{eq:ray-equation}
    x_b = p+s\frac{\mu - p}{||\mu - p||_2}.
\end{equation}
Thus, finding $x_b$ requires solving 
\begin{equation}
    (\mu - x_b)\Sigma^{-1}(\mu - x_b)=-2\log \tau
\end{equation}
for the scalar $s$, i.e. for each query pixel in the 2D gradient field. This is
done by parametrizing a quadratic equation, yielding the solution
\begin{align}
    s &= \| \mu-p \|_2 \left( 1 \pm \sqrt{\frac{-2\log\tau}{K}} \right) \\
    K &= a(\mu_x - p_x)^2 + 2b(\mu_x-p_x)(\mu_y-p_y) + c(\mu_y-p_y)^2,
\end{align}
with the coefficients $a,b,c$ from \cref{eq:conic}. The full derivation of the
solution is given in the supplementary material.}
A 1D illustration of the cut-off points and linear surrogate functions are
shown in \cref{fig:piecewise_grad_1D}.



The choice of slope $m$ influences the range of action of the truncated gradient
in the image plane: a lower value of $m$ yields a near-constant long-range
gradient approximating the cut-off gradient, and a higher value of $m$ yields a
sharp decay from the cut-off gradient. This cut-off gradient is defined as
$G^\prime(x)$, where $x$ is the solution to $G(x)=\tau$.

\begin{figure}[bt]
\centering
\includegraphics[width=0.9\columnwidth]{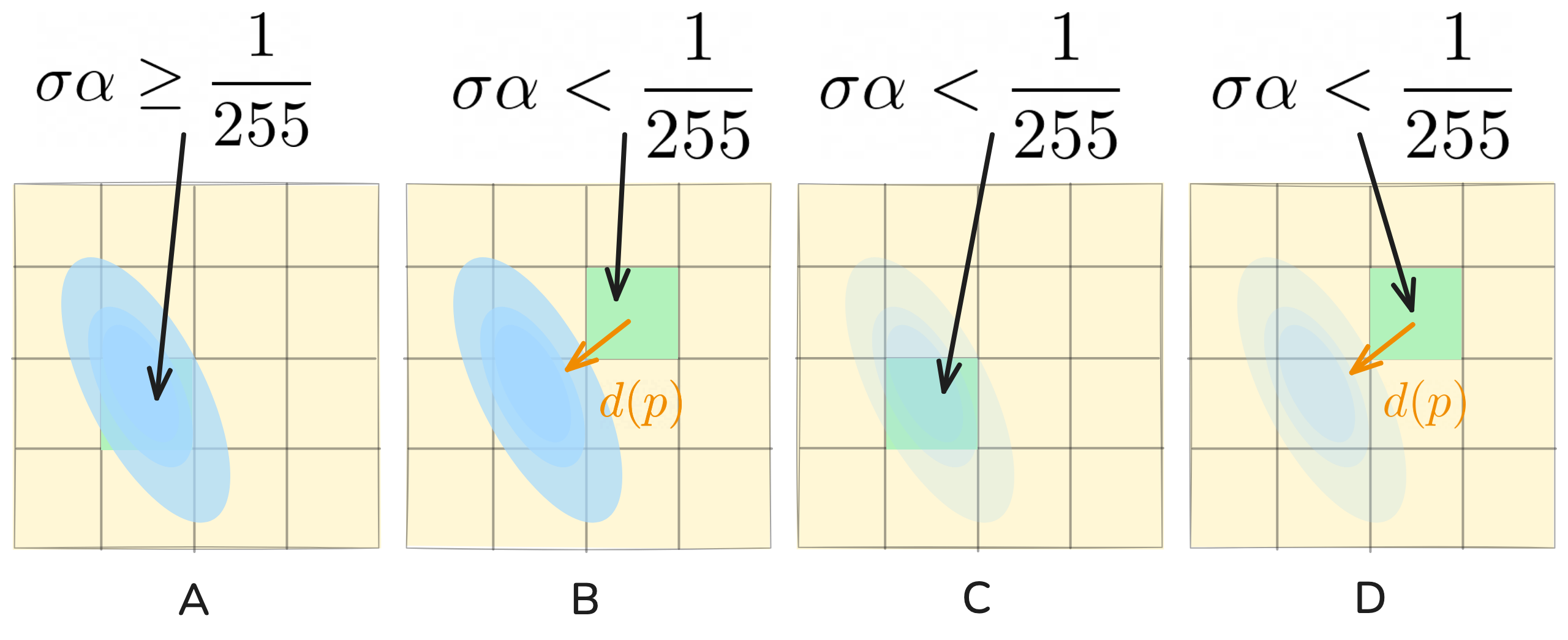}
\caption{Different conditions lead to different gradient paths for a given
Gaussian when looking at the error of a pixel in the tile. Depending on the
Gaussian density $\alpha$ at pixel $p$, the distance $d(p)$ to the isocontour of
density threshold $\tau$, pixel $p$ may contribute to the Gaussian
of opacity $\sigma$. A) The splat uses the
true gradient. B) The splat is filtered out and uses the truncated gradient,
with $d(p)>\tau$. C) The splat is filtered out and
receives no gradient, with $d(p)<\tau$. D) The splat
is filtered out and uses the truncated gradient, with $d(p)>\tau$.}
\label{fig:grad-paths}
\end{figure}

\paragraph*{Convergence analysis.}
When using the truncated gradient, there is a risk of model divergence when the
accumulated truncated gradient surpasses the true gradient.
This may arise when a Gaussian is a very bad fit, \textit{i.e.} it does not receive gradients
from the normal path, but also when it is a good fit, \textit{i.e.} its true gradient is near
zero at the local minimum (see \cref{fig:truncated-grad}).
Therefore, it is challenging to optimize poor fits well while maintaining the good fits reconstruction.

To prevent divergence, we found that the optimal strategy follows Occam's razor:
we only apply the truncated gradient defined in \cref{eq:dgdmu_truncated} for \emph{dead}
Gaussians, \emph{i.e.} $\sigma_i < \tau_\text{pruning}$. This means that
Gaussians whose opacity is below the pruning threshold would get pruned before
they can converge to distant pixels. To circumvent this, we employ an
interleaved strategy of optimization via the truncated gradient followed by
normal optimization with Adaptive Density Control (ADC) turned on. This has the
added benefit that perturbations to the accumulated viewspace gradient don't
affect ADC. In addition, we delay the pruning of primitives until the the end of
training.

\subsection{Implementation}
We integrate our truncated gradients into the backward pass of 3D Gaussian Splatting. 
In practice, we only assign the truncated gradient when the squared Mahalanobis
distance is greater than the cut-off threshold, since below that threshold we fallback
to the default case where the truncated gradient is not needed. In addition, we
only apply the truncated gradient when $\frac{\partial L}{\alpha_i} <0$,
\textit{i.e.} when the Gaussian isn't increasing the loss by alpha-blending,
otherwise we not only get a long-range pull but also a long-range push, which
causes Gaussians to repel each other. This results in three gradient update
paths for the pixel error with respect to the Gaussian 2D means (see
\cref{fig:grad-paths} for visual examples):
\begin{itemize}
    \item \textbf{Normal path}: Gaussians receive gradients from the normal
    path when they go through the high-pass filter, \textit{i.e.} $\sigma
    \alpha >= \frac{1}{255}$ (\cref{fig:grad-paths}).
    
    \item \textbf{Truncated path}: Gaussians receive gradients from the truncated path whenever they do not contribute to the pixel's rasterization,
    \textit{i.e.} $\sigma \alpha < \frac{1}{255}$ due to a low density $\alpha$ (see
    \cref{fig:grad-paths}B, or due to a low opacity $\sigma$ (see
    \cref{fig:grad-paths}C). In both cases, their Mahalanobis
    distance from the pixel is above the threshold. In the latter case, we only compute the gradient for the opacity $\sigma$. 
    \item \textbf{Skip path}: Gaussians do not receive any gradients when
    they do not contribute to the pixel, \textit{i.e.} $\sigma
    \alpha < \frac{1}{255}$, and their Mahalanobis distance from the pixel is below the threshold (\cref{fig:grad-paths}D).
\end{itemize}

Since we are only considering the gradient of the loss with respect to the
Gaussian mean, this modification results in a displacement of the Gaussian
whenever the accumulated truncated gradient magnitude is higher than the
accumulated true gradient coming from pixels, to which the Gaussian contributes.

Finally, to make use of the truncated gradient, we expand the tile-coverage radius
of each dead Gaussian whose radius is a below a threshold. This addresses the tile-based
culling of the rasterizer and enables long-distance pixel error contributions using
our truncated gradient field.

\paragraph*{Training performance.}
Without careful consideration, our truncated gradient method can be significantly
slower than the baseline in certain cases. To overcome this, we only apply
radius expansion and truncated gradient computation for \emph{dead} Gaussians.
In addition, we give dynamic methods a special treatment to reduce computation
by ignoring the contributions of static pixels, according to viewpoint-dependent
motion masks. We create these motion masks via frame-difference and image
processing on the input videos. With those optimizations, our method is
$1-10\times$ slower to train than the baseline, depending on the scene and
primitive count. At inference time however, the proposed modifications have no
effect on the rendering speed. \highlight{We discuss the performance overhead
in detail in the supplementary material.}
\section{Evaluation}
\begin{table*}[tb]
\centering

\caption{Quantitative comparison on static scene reconstruction benchmarks under different initialization settings.}
\vspace{0.5em}

\label{tab:main_static_results}

\resizebox{\textwidth}{!}{
\begin{tabular}{l|ccccc|ccccc}
\toprule

\multirow{2}{*}{Method}

& \multicolumn{5}{c|}{\textbf{\Ourbench{}}}
& \multicolumn{5}{c}{\textbf{Mip-NeRF360}} \\

& LPIPS$\downarrow$
& SSIM$\uparrow$
& PSNR$\uparrow$
& Train$\downarrow$
& \#GS$\downarrow$


& LPIPS$\downarrow$
& SSIM$\uparrow$
& PSNR$\uparrow$
& Train$\downarrow$
& \#GS$\downarrow$ \\

\midrule
\multicolumn{11}{c}{\textbf{Random initialization}} \\
\midrule



3DGS
& 0.2106 & 0.8351 & 24.74 & \textbf{14 min} & 1.126M 
& 0.3481 & 0.6718 & 20.92 & \textbf{27 min} & 1.180M \\  

\textbf{3DGS + Ours}
& \textbf{0.1974} & \textbf{0.8511} & \textbf{25.16} & 55 min & \textbf{0.799M} 
& \textbf{0.3343} & \textbf{0.6933} & \textbf{22.18} & 56 min & \textbf{0.971M} \\ 
\midrule

2DGS
& 0.4354 & 0.4960 & 18.80 & \textbf{23 min} & \textbf{1.231M} 
& 0.3943 & 0.6391 & 19.82 & \textbf{34 min} & 1.511M \\ 
\textbf{2DGS + Ours}
& \textbf{0.3883} & \textbf{0.6624} & \textbf{20.70} & 33 min & 1.446M 
& \textbf{0.3886} & \textbf{0.6547} & \textbf{20.33} & 55 min & \textbf{1.071M} \\  

\midrule
\multicolumn{11}{c}{\textbf{COLMAP initialization}} \\
\midrule

3DGS
& 0.1270 & 0.9058 & 30.07 & \textbf{9 min} & 1.217M 
& \textbf{0.2149} & \textbf{0.8212} & 27.70 & \textbf{21 min} & 2.496M \\ 

\textbf{3DGS + Ours}
& \textbf{0.1245} & \textbf{0.9097} & \textbf{30.86} & 43 min & \textbf{0.768M} 
& 0.2197 & 0.8205 & \textbf{27.84} & 86 min & \textbf{2.035M}\\  
\midrule

2DGS
& 0.2017 & 0.8378 & 25.00 & \textbf{23 min}  & 1.801M  
& 0.2342  & \textbf{0.8117} & 27.17 & \textbf{44 min} &  3.068M \\
\textbf{2DGS + Ours}
& \textbf{0.1962} & \textbf{0.8444} & \textbf{26.35} & 32 min  & \textbf{1.048M} 
& \textbf{0.2403} & 0.8104 & \textbf{27.33} & 80 min & \textbf{2.701M}  \\

\bottomrule
\end{tabular}
}
\end{table*}

To validate the effectiveness and generality of our proposed optimization framework, we evaluate our method on both static and dynamic scenes across indoor and outdoor environments. We compare against competing Gaussian splatting variants on public benchmarks as well as on our newly introduced dynamic dataset designed to exhibit large spatio-temporal divergence and challenging initialization conditions.

\subsection{Datasets \& metrics}

\paragraph*{Static datasets.}
In addition to still frames from our novel benchmark, we evaluate our method on the \emph{Mip-NeRF360 benchmark}~\cite{mipnerf360}. 
Mip-NeRF360 contains unbounded inward-facing indoor and outdoor scenes with large depth variation, detailed backgrounds, and 360-degree camera trajectories. The dataset is particularly challenging for Gaussian splatting methods due to its large-scale spatial extent and weakly constrained distant geometry.
For all datasets, we follow the evaluation protocol of Mip-NeRF360 and use every eighth image as part of the test set, while the remaining images are used for training.

\paragraph*{Dynamic datasets.}
Our novel synthetic dataset, \emph{\ourbench{}}, is composed of 6 scenes:
\emph{alley, windy tree,
water cup, neon city, bouncy balls, underwater}. Each scene is a 300-frame
sequence, from fluid simulation to hand-crafted animation, made and rendered
with Blender~\cite{blender}.
We rendered the scenes with camera rigs ranging from 25 to 45 cameras, arranged
in semi circles or full circles, and with a resolution of $1600\times 900$. All
scenes are rendered with path tracing at $30$fps, for a duration of $10$s each.
As we aim to raise the bar of dynamic Gaussian splatting methods, we consider
challenging scene dynamics with multiple novel view points, ranging from 1 to 4
test view points.

To show that our method performs well across all types of scenes, we also use
the Neural 3D Video benchmark~\cite{li2022neural}. It is composed of six $10$s
clips of indoor activities with $20$ cameras, including $1$ test view point.

\paragraph*{Metrics}
We evaluate rendered image quality using PSNR, SSIM, and LPIPS computed against the corresponding ground-truth images. PSNR measures pixel-wise reconstruction fidelity, SSIM evaluates structural similarity, and LPIPS measures perceptual similarity in feature space. All reported metrics are averaged across all test views.

\subsection{Experiments}

\paragraph*{Static scene reconstruction.} 
We compare our method against existing Gaussian splatting approaches on the Mip-NeRF360 benchmark and the static subset of our dataset, evaluating under both random initialization and COLMAP-based initialization. As shown in~\cref{tab:main_static_results}, our method consistently improves reconstruction quality across all settings and base methods (3DGS and 2DGS). The gains are particularly evident under random initialization, demonstrating stronger robustness to poor starting conditions, while improvements remain clear even with COLMAP initialization. Our method also tends to produce more compact representations with fewer Gaussians.

\paragraph*{Dynamic scene reconstruction.}
\begin{figure*}
    \centering
    \includegraphics[width=0.9\textwidth]{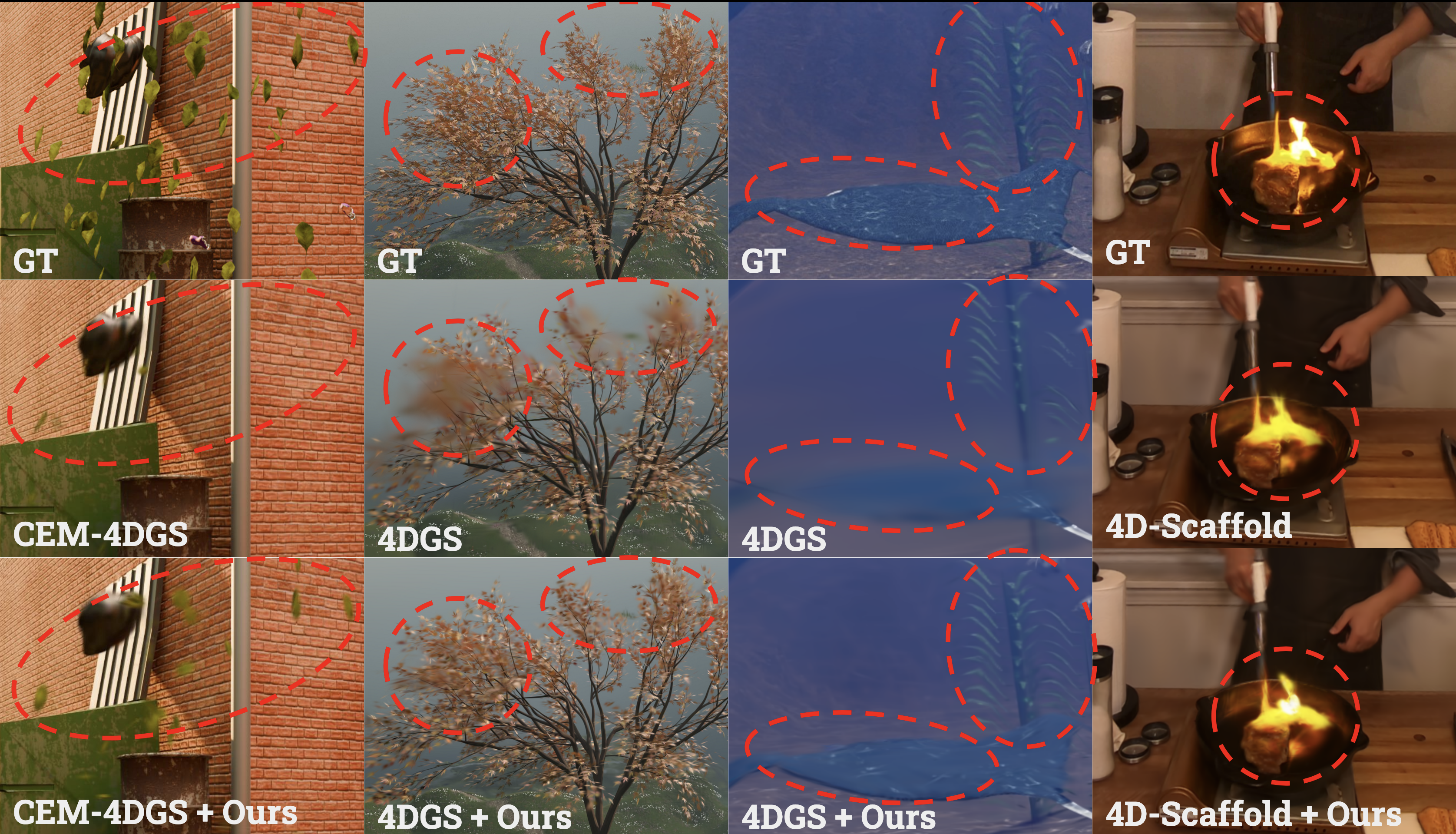}
    \caption{Qualitative comparison of our truncated gradient and the baselines
    on dynamic scene reconstruction benchmarks. From left to right:
    \emph{alley}, \emph{windy tree}, \emph{underwater} scenes of \ourbench{}, and
    on the last column the \emph{flame steak} scene of
    N3DV~\cite{li2022neural} }.
    \label{fig:main_dynamic}
\end{figure*}

\begin{figure*} \centering
    \includegraphics[width=0.99\textwidth]{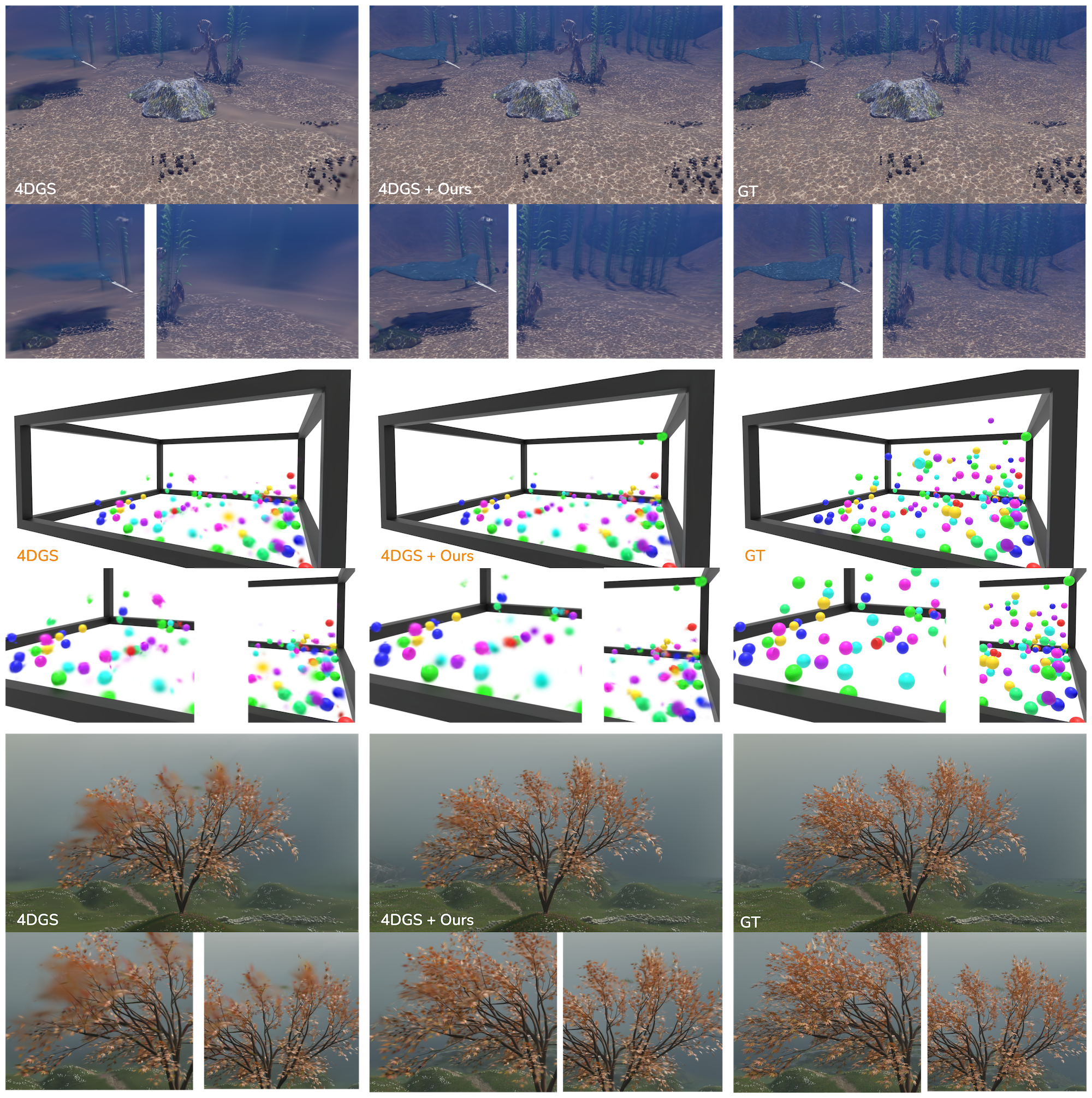}
    \caption{Qualitative comparison of our truncated gradient and the baselines
    on dynamic scene reconstruction benchmarks. From top to bottom:
    \emph{underwater}, \emph{bouncy balls} and \emph{windy tree}.}
    \label{fig:large_dynamic_ours_1}
\end{figure*}

\begin{figure*}
    \centering
    \includegraphics[width=0.99\textwidth]{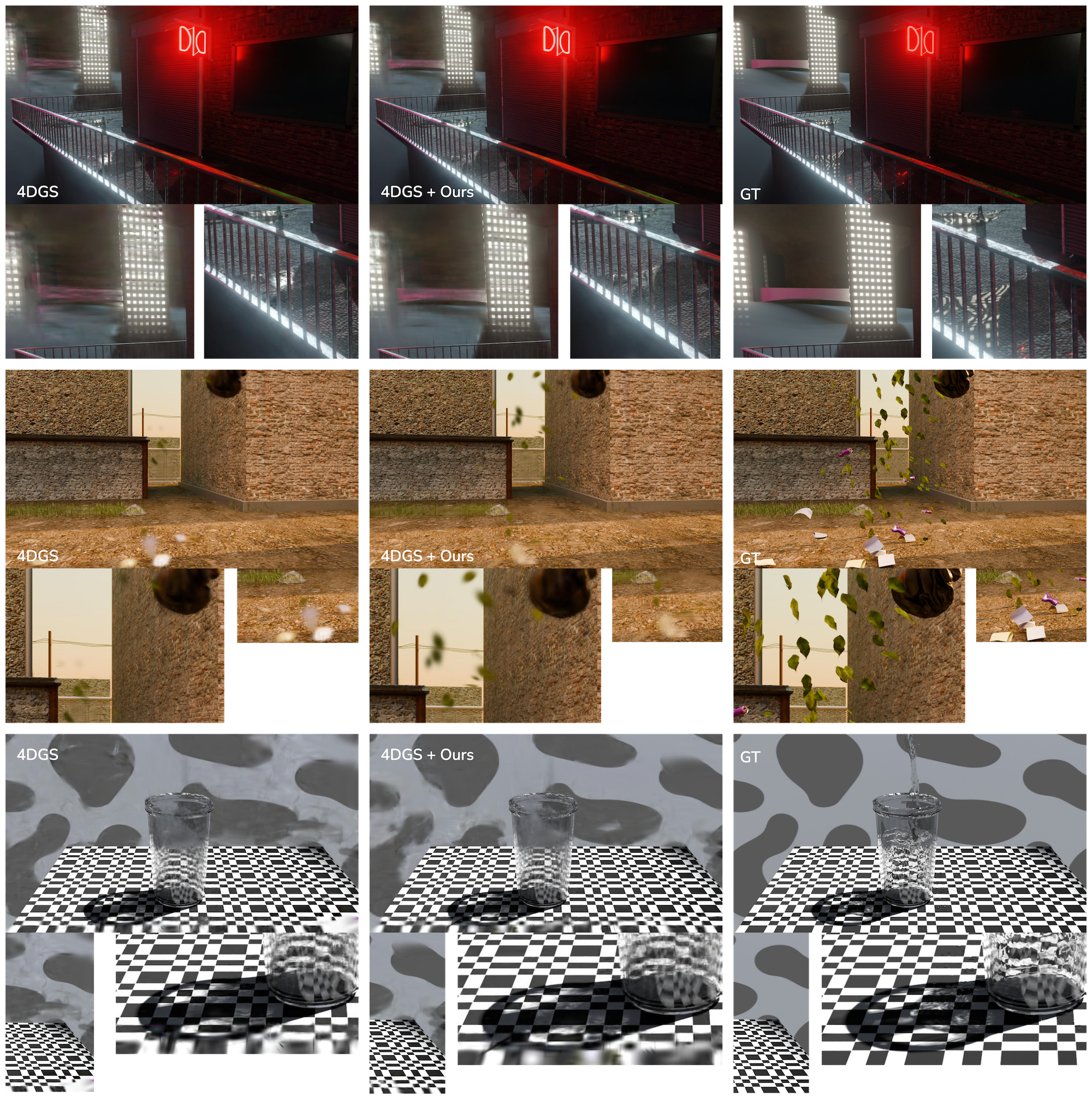}
    \caption{Qualitative comparison of our truncated gradient and the baselines
    on dynamic scene reconstruction benchmarks. From top to bottom:
    \emph{Neon city}, \emph{Alley} and \emph{Cup}.}
    \label{fig:large_dynamic_ours_2}
\end{figure*}

\begin{table}[t]
\centering
\caption{Quantitative comparison on the Neural 3D Video benchmark~\cite{li2022neural} for dynamic scene reconstruction.}

\vspace{0.5em}
\label{tab:dyn_n3dv_main}

\begin{tabular}{l|cccc}
\toprule

Method
& LPIPS$\downarrow$
& SSIM$\uparrow$
& PSNR$\uparrow$
& \#GS$\downarrow$ \\

\midrule

4DGS
& 0.1360 & 0.9443 & 30.22  & 2.806M \\

\textbf{4DGS + Ours}
& \textbf{0.1339} & \textbf{0.9493} & \textbf{31.70}  & \textbf{1.733M} \\
\midrule

CEM-4DGS
& \textbf{0.1328} & 0.9512 & 32.07  & \textbf{0.331M}  \\

\textbf{CEM-4DGS + Ours}
& 0.1359 & \textbf{0.9519} & \textbf{32.29}  & 0.352M  \\

\midrule
4D-Scaffold
& \textbf{0.1286} & \textbf{0.9502} & 31.83  & 0.583M \\

\textbf{4D-Scaffold + Ours}
& 0.1311 & 0.9487 & \textbf{31.94} &  \textbf{0.556M}  \\

\bottomrule
\end{tabular}
\end{table}

\begin{table}[t]
\centering
\caption{Quantitative comparison on our novel benchmark for dynamic scene reconstruction. 
$\phi$ and $\theta$ refer to sparse and dense COLMAP reconstructions used as initialization, respectively.}
\vspace{0.5em}
\label{tab:dync_ours_main}
\begin{tabular}{l|cccc}
\toprule
Method & LPIPS$\downarrow$ & SSIM$\uparrow$ & PSNR$\uparrow$ & \#GS$\downarrow$ \\
\midrule
4DGS ($\phi$) & 0.2326 & 0.8329 & 25.47 & 2.409M \\
\textbf{4DGS ($\phi$) + Ours} & \textbf{0.2035} & \textbf{0.8574} & \textbf{26.08} & \textbf{1.715M} \\
\midrule
4DGS ($\theta$) & 0.2288 & 0.8224 & 26.02 & 2.970M \\
\textbf{4DGS ($\theta$) + Ours} & \textbf{0.1496} & \textbf{0.8923} & \textbf{27.56} & \textbf{2.333M} \\
\midrule
CEM-4DGS & 0.2767 & \textbf{0.7855} & \textbf{22.80} & \textbf{0.704M} \\
\textbf{CEM-4DGS + Ours} & \textbf{0.2670} & 0.7845 & 22.63 & 0.824M \\
\bottomrule
\end{tabular}
\end{table}

\begin{table*}[t]
\centering
\caption{Ablation study of key components on the \emph{alley} and \emph{tree} scenes of \ourbench{}.}
\vspace{0.5em}
\label{tab:ablation_components}

\begin{tabular}{l|ccc|ccc}
\toprule
& \multicolumn{3}{c|}{\emph{Alley}} &
\multicolumn{3}{c}{\emph{Tree}} \\
Configuration &
LPIPS$\downarrow$ &
SSIM$\uparrow$ &
PSNR$\uparrow$ &
LPIPS$\downarrow$ &
SSIM$\uparrow$ &
PSNR$\uparrow$ \\
\midrule

w/o truncated gradient
& 0.1350 & 0.8891 & 28.82
& \highlight{0.1789} & \highlight{0.8681} & \highlight{28.03} \\

w/o radius padding
& \textbf{0.1289} & \underline{0.8913} & 28.83
& \highlight{0.1772} & \highlight{0.8694} & \highlight{28.42} \\


w/o delayed pruning
& 0.1309 & 0.8909 & 28.82
& \highlight{0.1707} & \highlight{\textbf{0.8753}} & \highlight{\textbf{28.79}} \\

\highlight{w/o dead Gaussians filtering}
& \highlight{0.3716}
& \highlight{0.6153}
& \highlight{21.87}
& \highlight{0.1755}
& \highlight{0.8728}
& \highlight{28.67} \\

\highlight{w/o negative alpha filtering}
& \highlight{\underline{0.1291}}
& \highlight{\textbf{0.8922}}
& \highlight{\textbf{29.14}}
& \highlight{\underline{0.1701}}
& \highlight{\underline{0.8733}}
& \highlight{\underline{28.70}} \\

\midrule

Full method
& \textbf{0.1289}
& 0.8887
& \underline{29.11}
& \highlight{\textbf{0.1685}}
& \highlight{0.8714}
& \highlight{28.58} \\

\bottomrule
\end{tabular}
\end{table*}





We evaluate our approach on the Neural 3D Video benchmark and our novel synthetic dataset using popular dynamic baselines. All dynamic sequences are initialized from the first frame via COLMAP reconstruction and optimized sequentially over time. This setup progressively increases optimization difficulty due to accumulating temporal displacements. Our truncated gradient formulation is applied on top of the baselines with minimal modifications, highlighting its plug-and-play compatibility and effectiveness in handling large spatio-temporal divergence. 

Our method delivers consistent improvements when integrated with 4DGS~\cite{4dgs}, CEM-4DGS~\cite{kang2025cem4dgs}, and 4D-Scaffold~\cite{cho2024scaffold} (\cref{tab:dyn_n3dv_main} and \cref{tab:dync_ours_main}). Qualitative comparisons (\cref{fig:main_dynamic}, \cref{fig:large_dynamic_ours_1} and \cref{fig:large_dynamic_ours_2}) show better recovery of complex motion, fluid effects, and fine geometric details where baselines exhibit floaters, blurring, or missing structures.

Overall, these experiments demonstrate that our piecewise truncated gradient approach provides generalizable and robust improvements for both static and dynamic 3D Gaussian Splatting across different initializations and scene types.

\subsection{Ablation Study}

\paragraph*{Ablation on optimization design choices.}

Our method introduces three key components: the truncated gradient, radius padding, and delayed pruning. Since these components are implemented directly within the Gaussian rasterization and optimization pipeline, their behavior is independent of whether the scene is static or dynamic. To isolate their effects while keeping computational cost manageable, we perform all ablations on the alley scene of \ourbench{}.

\cref{tab:ablation_components} summarizes the contribution of each component. Removing the truncated gradient causes the largest performance drop, confirming its critical role in addressing vanishing gradients and enabling effective long-range movement of primitives across large displacements in dynamic scenes. Disabling radius padding yields competitive but inferior results, as padding ensures dead Gaussians are assigned to relevant tiles, enabling the truncated gradient to take effect over longer distances. Similarly, early pruning of low-opacity Gaussians prevents them from receiving long-range gradients. Keeping them until the end allows recovery and improves final reconstruction. \highlight{Applying the truncated gradient to all Gaussians degrades performance, as the optimization bias introduced by the surrogate linear model can cause training divergence even for active Gaussians. Without filtering by the sign of the alpha gradient, we not only get a long-range pull but also a long-range push, which causes Gaussians to repel each other.} The full method achieves the best performance by combining stronger gradient flow, sufficient tile coverage, and preserved exploration capacity.

\subsection{Discussion and Limitations}

Our method improves optimization robustness by enlarging the effective support region of Gaussian primitives, allowing poorly initialized or weakly constrained Gaussians to receive meaningful gradients from a larger set of pixels. However, this increased optimization coverage comes at the cost of additional computation, as more Gaussian-pixel pairs participate in the backward pass. As a result, our method generally requires longer training times than the original rasterization pipeline. This reflects a fundamental trade-off between optimization efficiency and reconstruction robustness: improved gradient propagation leads to better scene reconstruction quality, but incurs additional computational overhead.

We note that, except for the delayed pruning schedule, we did not retune the original densification and pruning hyperparameters of the baseline methods. This conservative integration strategy ensures broad compatibility with existing 3DGS frameworks but leaves room for further gains, particularly in metrics such as SSIM that are sensitive to densification strategies. Exploring joint optimization of our truncated gradients with tailored densification/pruning policies is a promising direction for future work.
\section{Conclusion}

We presented TruncGradGS, an optimization framework for improving Gaussian splatting through piecewise truncated gradient updates. By replacing the near-zero tails of the Gaussian derivative with a continuous surrogate, our method enlarges the effective optimization support of Gaussian primitives and alleviates the vanishing-gradient problem inherent to tiled Gaussian rasterization. Extensive experiments on both static and dynamic scene reconstruction benchmarks demonstrate consistent improvements in reconstruction quality and robustness across different initialization settings and Gaussian representations. In addition, we introduced a new synthetic benchmark for dynamic Gaussian splatting containing challenging scenes with large spatio-temporal divergence. Our results suggest that improving gradient propagation is an effective and complementary direction for advancing Gaussian-based scene reconstruction methods.

\paragraph*{Acknowledgment.}
This project is supported by Research Ireland under the Research Ireland Frontiers for the Future Programme - Project, award number 22/FFP-P/11522. The team gratefully acknowledges Adam Handschuh for creating the dataset, Leon Andorfi for supporting the experiments, and Dolby Laboratories, Inc. for their gift fund. 

\bibliographystyle{eg-alpha-doi} 
\bibliography{egbib}       


\end{document}


\title{TruncGradGS: Improved 3D Gaussian Splatting via Truncated Gradient Updates -- Supplementary Material}

\author[T. Morales, N.-Q. Le-Pham, R. Atkins \& B.-S. Hua]
{\parbox{\textwidth}{\centering
Théo Morales$^{1}$\orcid{0000-0002-2275-0895}
\quad
Nhat-Quynh Le-Pham$^{1}$\orcid{0000-0002-8668-9691}
\quad
Robin Atkins$^{2}$
\quad
Binh-Son Hua$^{1}$\orcid{0000-0002-5706-8634}
\\[2mm]
{\parbox{\textwidth}{\centering
$^1$Trinity College Dublin \quad $^2$Dolby Laboratories
}
}}
}
\maketitle

In this supplementary material, we provide additional details on our method and 
results on each scene in the datasets. 
Please see our supplementary videos for the visual results of the scenes.

\paragraph*{Full derivation of the boundary point solution.}
\highlight{We here provide the full derivation for the solution of
\begin{equation} \label{eq:solution-point}
    x_b = p+s\frac{\mu - p}{\| \mu - p \|_2}.
\end{equation}
As stated in Section 3.3., finding $x_b$ requires solving 
\begin{equation}
    (\mu - x_b)\Sigma^{-1}(\mu - x_b)=-2\log \tau
\end{equation}
for the scalar $s$, i.e. for each query pixel in the 2D gradient field. This can
be written in conic form with coefficients $a,b,c$ (see Equation 8 in Section
3.3.) as
\begin{equation} \label{eq:isocontour-conic}
   ad_x^2 + 2bd_xd_y+cd_y^2=l \quad \textrm{with} \quad l=-2\log \tau.
\end{equation}}

\highlight{First, we introduce the distance-space formulation $d=\mu-p$ used
in the original rasterizer implementation. In this space, the solution
$x_b$ gives $d^*=\mu-x_b$, or after substituting \cref{eq:solution-point}:
\begin{equation}
    d^*=\mu-(p+s\frac{\mu - p}{||\mu - p||_2}).
\end{equation}
We can simplify this and decompose each variable into
\begin{align}
    d_x^* &= u - s \cdot \frac{u}{q} \quad \textrm{with} \quad u = \mu_x - p_x, \\
    d_y^* &= v - s \cdot \frac{v}{q} \quad \textrm{with} \quad v = \mu_y - p_y,
\end{align}
and with $q=||\mu-p||_2$.
We can now reformulate \cref{eq:isocontour-conic} as
\begin{equation}
   a{d_x^*}^2 + 2bd_x^*d_y^*+c{d_y^*}^2 -l=0
\end{equation}
with 
\begin{align}
    d_x^*&=u(1-\frac{s}{q}) \\
    d_y^*&=v(1-\frac{s}{q}),
\end{align}
and after substitution:
\begin{equation}
    (1-\frac{s}{q})^2 [a\mu^2+2buv+cv^2]-l=0.
\end{equation}
Define $K=au^2+2buv+cv^2$; then
\begin{align}
    (1-\frac{s}{q})^2 K-l=0 \equiv (1-\frac{s}{q})^2 &= \frac{l}{K} \\
    1-\frac{s}{q} &= \pm \sqrt{\frac{l}{K}} \\
    1 \pm \sqrt{\frac{l}{K}} &= \frac{s}{q} \\
    s &= q(1 \pm \sqrt{\frac{l}{K}}).
\end{align}
This is equivalent to
\begin{align}
    s &= \| \mu-p \|_2 \left( 1\pm \sqrt{\frac{-2\log\tau}{K}} \right) \\
    K &= a(\mu_x - p_x)^2 + 2b(\mu_x-p_x)(\mu_y-p_y) + c(\mu_y-p_y)^2,
\end{align} 
as written in Section 3.3. As expected, there are two solutions because the ray intersect
the 2D splat at the entry and exit points. We naturally only consider the entry point, 
\emph{i.e.} the minimum of the two solutions corresponding to the smallest ray distance.}

\paragraph*{Details of hyperparameters and schedules.}
\highlight{For the density threshold, we use the original discard threshold of 1/255. We fine-tuned the slope to 1e-7, the radius padding to 96px, and the opacity threshold for dead Gaussian selection to 0.01. Regarding the training scheme, we start with adaptive density control (ADC) turned on and switch it off after 3150 iterations for a truncated gradient phase of 5000 iterations. This interleaves ADC and our modified backward pass without ADC, until iteration 25,000 where we only use our modified backward pass.
For our motion masks, refer to section 3.4. }

\paragraph*{Computational cost.}
\highlight{
Motion masks are only used in the dynamic setting. Without motion masks, the baseline 4DGS exceeds the available GPU memory budget and cannot be trained under our experimental setup. Therefore, motion masks are necessary to make dynamic-scene training feasible.}

\highlight{
To quantify the computational overhead of each component, we report the average GPU memory usage and iteration time measured during the ablation experiments in Table~5 of our main paper. The average GPU memory consumption is 10.93GB for the full method, compared to 13.07GB without truncated gradients, 9.19GB without radius padding, and 11.09GB without delayed pruning.}

\highlight{
With adaptive density control (ADC) enabled, the average iteration time is 34ms for the full method, compared to 44ms without truncated gradients, 47ms without radius padding, and 46ms without delayed pruning. Since ADC introduces additional computational overhead, we also report timings with ADC disabled while keeping truncated gradients enabled. In this setting, the average iteration time is 120ms for the full method, compared to 56ms without truncated gradients, 58ms without radius padding, and 73ms without delayed pruning.}

\highlight{
These results indicate that truncated gradients improve both memory efficiency and training speed when ADC is enabled, while radius padding and delayed pruning introduce only modest computational overhead relative to the overall training process.
}

\section{Conclusion}

We presented TruncGradGS, an optimization framework for improving Gaussian splatting through piecewise truncated gradient updates. By replacing the near-zero tails of the Gaussian derivative with a continuous surrogate, our method enlarges the effective optimization support of Gaussian primitives and alleviates the vanishing-gradient problem inherent to tiled Gaussian rasterization. Extensive experiments on both static and dynamic scene reconstruction benchmarks demonstrate consistent improvements in reconstruction quality and robustness across different initialization settings and Gaussian representations. In addition, we introduced a new synthetic benchmark for dynamic Gaussian splatting containing challenging scenes with large spatio-temporal divergence. Our results suggest that improving gradient propagation is an effective and complementary direction for advancing Gaussian-based scene reconstruction methods.

\paragraph*{Acknowledgment.}
This project is supported by Research Ireland under the Research Ireland Frontiers for the Future Programme - Project, award number 22/FFP-P/11522. The team gratefully acknowledges Adam Handschuh for creating the dataset, Leon Andorfi for supporting the experiments, and Dolby Laboratories, Inc. for their gift fund.

\paragraph*{Adaptation to other baselines.}
\highlight{The flexibility of our method enables the modification of any baseline
method with our truncated gradient and training schedule. Our modifications are
valid for any 3DGS-based method; they are twofold: \emph{backward pass} and \emph{training schedule}.}

\highlight{For the backward pass, we modify the rasterizer in several places.
First, we add padding to the radius of the bounding box of a Gaussian, if its
opacity is below the threshold. This is done in the forward pass, but due to
alpha filtering, it has no effect on rasterization and is only used for the
backward pass. Second, we add a branch to the alpha filtering of the backward
pass, and compute the truncated gradient, which accumulates in the 2D means
gradient tensor. In this branch, we add the negative alpha guard: only positive
alpha gradients result in truncated gradient accumulation.}

\highlight{For the training schedule, all methods that utilize the traditional
Adaptive Density Control (ADC) schedule can be modified the same way by
interleaving ADC phases with truncated gradient phases, as explained above. The
core principle is to turn off ADC until the loss has converged to some minimum,
before turning it on again without using the modified backward pass. This is
based on our observation that ADC interferes with our modified backward pass
with the truncated gradient. Therefore, this schedule may be tuned or adapted
for baselines which do not rely on the conventional ADC, i.e. MCMC-based density
control.}

\bibliographystyle{eg-alpha-doi} 
\bibliography{egbib}